\pdfoutput=1

\documentclass[11pt]{article}

\usepackage{acl}

\usepackage{times}
\usepackage{latexsym}
\usepackage{graphicx}
\usepackage{amsmath}
\usepackage{hyperref}
\usepackage{booktabs} 
\usepackage{multirow} 
\usepackage{multicol} 
\usepackage{subfig}
\usepackage{CJKutf8}
\usepackage{algorithm}
\usepackage{algorithmic}
\usepackage{tabularx}

\usepackage[T1]{fontenc}

\usepackage[utf8]{inputenc}

\usepackage{microtype}

\usepackage{inconsolata}

\title{GIELLM: Japanese General Information Extraction Large Language Model Utilizing Mutual Reinforcement Effect}

\author{
  Chengguang Gan\textsuperscript{1} \quad
  Qinghao Zhang\textsuperscript{2} \quad
  Tatsunori Mori\textsuperscript{1} \\
  \textsuperscript{1}Yokohama National University, Japan \\
  \texttt{gan-chengguan-pw@ynu.jp, tmori@ynu.ac.jp} \\ynu.jp
  \textsuperscript{2}Department of Information Convergence Engineering, \\
  Pusan National University, South Korea \\
  \texttt{zhangqinghao@pusan.ac.kr}
}

\begin{document}
\maketitle
\begin{abstract}
Information Extraction (IE) stands as a cornerstone in natural language processing, traditionally segmented into distinct sub-tasks. The advent of Large Language Models (LLMs) heralds a paradigm shift, suggesting the feasibility of a singular model addressing multiple IE subtasks. In this vein, we introduce the General Information Extraction Large Language Model (GIELLM), which integrates text Classification, Sentiment Analysis, Named Entity Recognition, Relation Extraction, and Event Extraction using a uniform input-output schema. This innovation marks the first instance of a model simultaneously handling such a diverse array of IE subtasks. Notably, the GIELLM leverages the Mutual Reinforcement Effect (MRE), enhancing performance in integrated tasks compared to their isolated counterparts. Our experiments demonstrate State-of-the-Art (SOTA) results in five out of six Japanese mixed datasets, significantly surpassing GPT-3.5-Turbo. Further, an independent evaluation using the novel Text Classification Relation and Event Extraction(TCREE) dataset corroborates the synergistic advantages of MRE in text and word classification. This breakthrough paves the way for most IE subtasks to be subsumed under a singular LLM framework. Specialized fine-tune task-specific models are no longer needed.

\end{abstract}

\section{Introduction}

The advent of Large Language Models (LLMs), including notable examples such as ChatGPT\citep{ouyang2022training}, GPT-4\citep{openai2023gpt4}, and LLaMA\citep{touvron2023llama}, has significantly transformed the landscape of Natural Language Processing (NLP). These models have showcased remarkable generalization capabilities, primarily attributed to their extensive pre-training on diverse datasets. This foundational knowledge equips LLMs to adeptly navigate a variety of NLP tasks, predominantly through a question-and-answer format that transforms inputs into coherent outputs.

\begin{figure}[!t]
\centering
\includegraphics[width=219 pt]{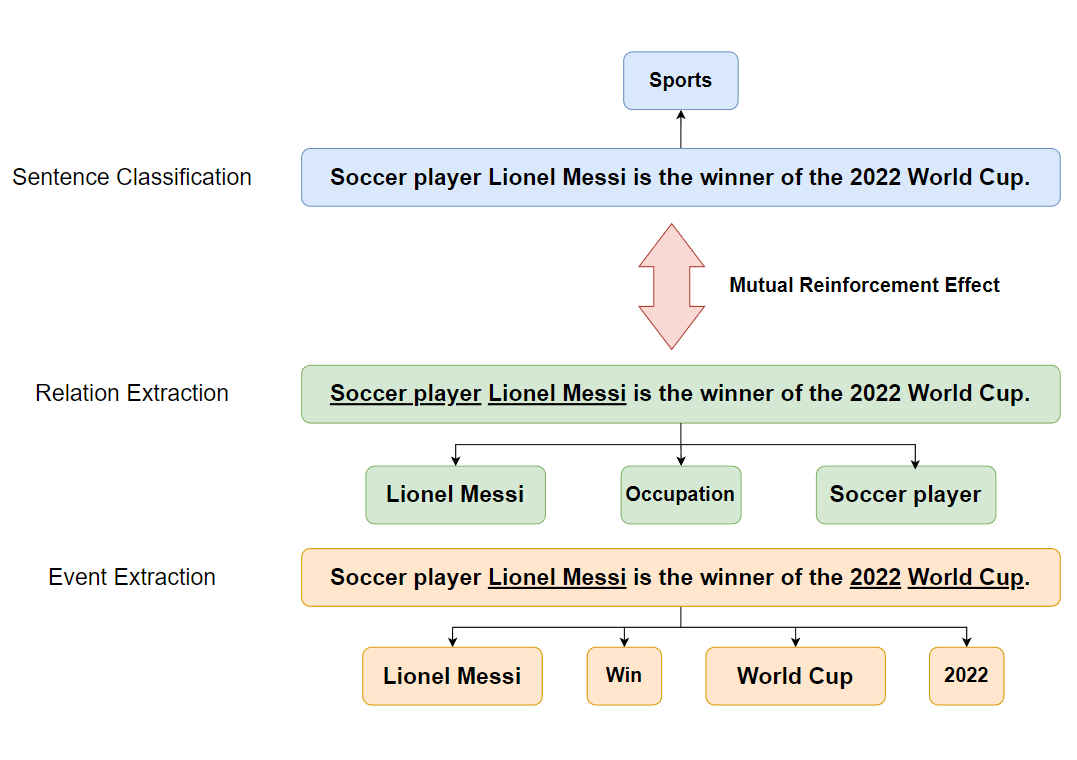}
\caption{\label{1figure1}The illustration depicts the Mutual Reinfocement Effect between the labels of Word-level Relation/Event Extraction and text-level Sentence Classification (SC) within a same text.}

\end{figure}

Despite their versatility, the primary application of mainstream LLMs has been centered around Q\&A-based chat functionalities, producing largely unstructured textual outputs. This approach, however, does not directly address the numerous NLP tasks that necessitate structured textual outputs, a critical requirement in Information Extraction (IE). IE encompasses a spectrum of sub-tasks, such as Text Classification\citep{lai2015recurrent}\citep{kowsari2019text}, Sentiment Analysis\citep{medhat2014sentiment}\citep{liu2022sentiment}, Named Entity Recognition (NER)\citep{nadeau2007survey}\citep{sang2003introduction}\citep{lample2016neural}, Relation Extraction (RE)\citep{mintz2009distant}\citep{etzioni2008open}, and Event Extraction (EE)\citep{8918013}, each demanding a unique output format.

Although recent efforts have explored the application of LLMs to these IE subtasks through In-context Learning\citep{brown2020language}\citep{dong2023survey} and Instruction Learning\citep{wei2021finetuned}\citep{dong2023survey}, there remains an absence of a cohesive, unified model and format capable of concurrently addressing all these IE sub-tasks. This gap highlights the need for further innovation in the field, aiming to develop an integrated framework that leverages the strengths of LLMs for comprehensive and structured information extraction.

In this research, we explore the application of LLMs to IE tasks, specifically focusing on generating structured text outputs. The extensive pre-training knowledge embedded in LLMs forms the foundation of our proposed Generalized Large Language Model for Information Extraction. A key innovation of our approach is the standardization of input and output formats, enabling the model to adeptly process all of IE subtasks. 

Distinctively, our methodology encompasses the use of a Mixed dataset for both training and testing phases. This approach entails dual-level classification - both word-level and text-level - on the same textual data, marking a significant departure from traditional IE research paradigms. This dual-level classification strategy is driven by the concept of Mutual Reinforcement Effect (MRE). According to \citet{10.1007/978-3-031-35320-8_18} research, there exists a potent synergistic relationship between word-level and text-level classification tasks. This interconnection is leveraged to concurrently enhance the model's proficiency in both tasks.

An illustrative example of this approach, involving Sentence Classification, Relation Extraction, and Event Extraction tasks, is presented in Figure \ref{1figure1}. Consider a scenario where the model identifies a given news text as sports-related. This initial categorization naturally leads to the extraction and classification of sports-related entities and events within the same text. For instance, the model might focus on segments mentioning sports figures like Lionel Messi or terms associated with sports professions like 'Soccer player.' The occurrence of these sports-related terms in the text reinforces the initial classification of the text as sports news. Conversely, the extraction and categorization of sports-related nouns and personalities further affirm the likelihood of the text being sports-oriented. This bidirectional enhancement epitomizes the essence of MRE, a central concept underpinning our research. 

In summary, our approach seeks to leverage the MRE to foster a more holistic understanding of text through simultaneous word-level and text-level comprehension, representing a novel and promising direction in IE research. The contributions of this paper are threefold:
\begin{enumerate}
\item We introduce the Format Converter, a novel approach that standardizes input and output formats across all Information Extraction (IE) subtasks, thus enabling a single model to efficiently handle diverse tasks.
\item We have developed the Sentence Classification Relation and Event Extraction (SCREE) dataset. Further, our ablation experiments on this dataset reaffirm the MRE phenomenon in the word-level and text-level classification tasks.
\item We have fine-tuned a Japanese General Information Extraction Large Language Model (GIELLM) using multiple mix datasets. This generalized model paves the way for future researchers to explore and expand upon Japanese IE more effectively.
\end{enumerate}

\section{Related Work}
\textbf{Universal Information Extraction.} Prior to the advent of LLMs, significant efforts were made to consolidate various IE subtasks into a unified model using a generative approach. Notably, UIE and USM stand out as pioneering frameworks, both employing a common strategy to integrate the input-output processes of IE subtasks. Specifically, UIE\citep{lu-etal-2022-unified} and USM\citep{lou2023universal} utilize T5\citep{raffel2020exploring} and RoBERTa\citep{liu2019roberta} models, respectively, for training, focusing solely on word-level IE challenges. In the \citet{yan-etal-2021-unified-generative}, the approach leverages a seq2seq framework to address flat, nested, and discontinuous NER tasks. GenIE\citep{josifoski-etal-2022-genie} employs a transformer-based model to extract information from unstructured text, applying global structural constraints. Similarly, InstructionNER\citep{wang2022instructionner} utilizes the T5 model to identify entity spans based on instructions and options. However, these methods are confined to NER, RE, and EE within IE. UniSA\citep{li2023unisa} innovatively combines a transformer decoder with a multimodal transformer encoder, using contrastive learning to adeptly handle various sentiment analysis subtasks within a singular model framework. The SLG Framework(Sentence-to-label Generation Framework)\citep{10.1007/978-3-031-35320-8_18} presents MRE for the first time. And uses the T5 model to process the Mix task for SC and NER. The experimental results show that the combined two tasks work better than individually.

\begin{figure*}[ht]
\centering
\includegraphics[width=440 pt]{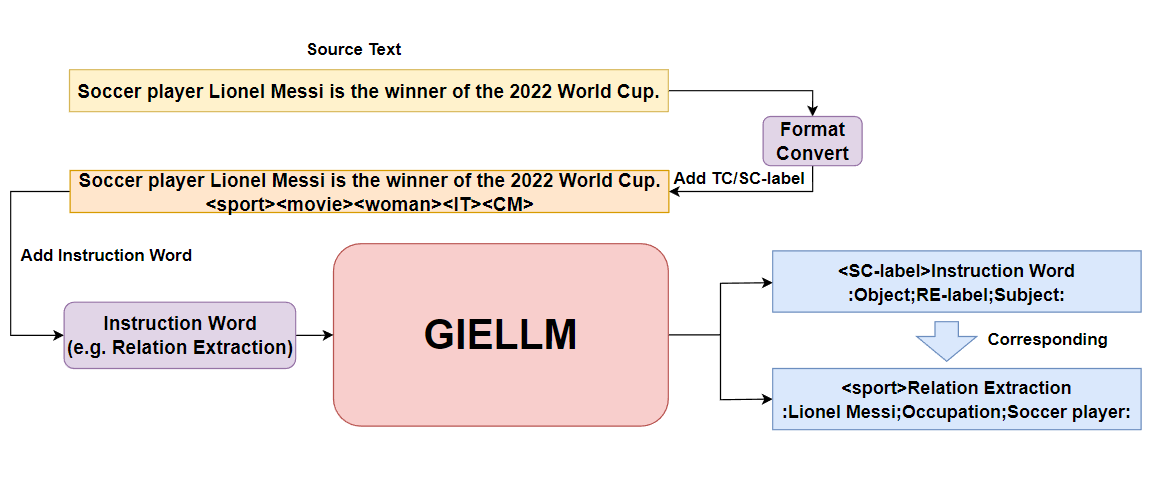}
\caption{\label{2figure2}The illustration depicts the GIELLM how the source text is processed and then output to GIELLM. Finally, the corresponding extraction information is obtained according to different IE subtasks.}

\end{figure*}

\textbf{LLMs Information Extraction.} After the advent of LLMs, basically all IE methods shifted to interacting with LLMs via prompts. Let the model answer the information that wants to extract. Code4UIE\citep{guo2023retrievalaugmented} innovatively employs a two-stage prompting process that inputs IE subtasks into LLMs in a standardized code format, subsequently eliciting the extraction of information in the same format. InstructUIE\citep{wang2023instructuie}, on the other hand, tailors specific instructional texts to different IE subtasks, guiding LLMs to output information in the required format. Diverging from the sole reliance on prompts and LLMs for IE subtasks, UniversalNER\citep{zhou2023universalner} leverages LLMs with capacities of 7B and 13B for NER. This approach involves constructing prompts to distill knowledge from the ChatGPT model, which is then used to fine-tune the 7B and 13B LLMs. Remarkably, this method has shown excellent performance in both in-domain and out-of-domain NER tasks. The USA model\citep{gan2023usa} introduces an innovative approach, the MRE, which intertwines word-level and sentence-level analysis in sentiment analysis tasks. This model, with a 7B capacity, adeptly handles complex subtasks of Sentiment Analysis, encompassing both text sentiment classification and part-of-speech sentiment classification.

However, the majority of these studies focused solely on word-level information extraction, neglecting text-level considerations. The MRE between these levels was largely overlooked, with only a minority of studies incorporating MRE. This research, in contrast, innovatively proposes employing a singular model to encompass all key IE subtasks, including NER, RE, EE, Sentiment Analysis, and Text Classification.

\section{Japanese General Information Extraction Large Language Model}

The fundamental aspect of IE lies in its classification task. This task can be stratified into two levels based on the unit of categorization: word-level and text-level. The core principle of MRE involves enabling the model to concurrently categorize a single sentence at both these levels. This approach fosters a synergistic enhancement, allowing the models to mutually reinforce their comprehension of both tasks. Building on this concept, we have developed and trained the GIELLM.

\begin{figure*}[!ht]
\centering
\includegraphics[width=440 pt]{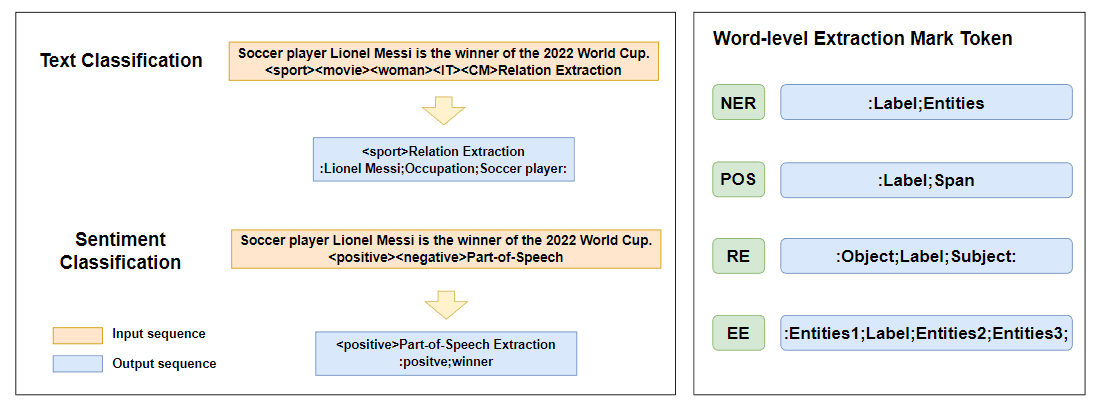}
\caption{\label{3figure3}The illustration depicts Examples of the Format Converter in different IE subtasks. Including Word-level and Text-level extracted information Mark Token.}

\end{figure*}

Figure \ref{2figure2} illustrates the comprehensive workflow of GIELLM as applied to information extraction. To capture the interaction between label schema and text, we define a text sequence for extraction as \( X \). This sequence contains \( n \) tokens: \( (X_1, X_2, \ldots, X_n) \). 
The sequence \( X \) is processed through a Format Converter, which labels words at the text level based on Text Classification (TC) or Sentiment Classification (SC). We denote the label word sequence for a TC task with \( n \) tokens as \( L = (L_1, L_2, \ldots, L_n) \).
We concatenate the text sequence \( X \) with the label sequence \( L \) to form a new sequence \( F = (X_1, X_2, \ldots, X_n, L_1, L_2, \ldots, L_n) \).
\begin{equation}
    F = (X_1, X_2, \ldots, X_n) \oplus (L_1, L_2, \ldots, L_n)
\end{equation}
This sequence is then processed by the Instruction Word (IW), creating the final input sequence \( I = (X_1, X_2, \ldots, X_n, L_1, L_2, \ldots, L_n, IW) \).
\begin{equation}
    I = (X_1, X_2, \ldots, X_n, L_1, L_2, \ldots, L_n,) \oplus (IW)
\end{equation}
Lastly, \( I \) is fed into the GIELLM to produce the output sequence \( O \), which includes the text classification label and the extracted information sequence (e.g. <sport>Relation Extraction:Lionel Messi;Occupation;Soccer player:).
\begin{equation}
    O = \textbf{GIELLM}(I)
\end{equation}
In the final formatted output, specific marks are visible, whose functions and meanings are thoroughly expounded in Section \ref{Format Converter}, "Format Converter." Additionally, the mechanisms underlying the model's ability to generate formatted content autonomously, without external directives, are comprehensively detailed in Section \ref{Evaluation}, "Training the GIELLM." The methodology for assigning appropriate Instruction Words (IW) for various Information Extraction (IE) subtasks is elucidated in Section \ref{Instruction Word}, "Instruction Word."

This research encompasses an extensive series of data preprocessing steps. All IE subtasks undergo standardization into a uniform format suitable for processing by the Large Language Model (LLM). Subsequently, the trained GIELLM produces a sequence of extracted information, consistently formatted to include both word-level and text-level extraction information. This output encapsulates the overall classification label of the text and detailed elements such as labels, entities, spans, and other relevant extracted information pertaining to the respective IE subtasks.

\subsection{Format Converter}\label{Format Converter}

In all general generative IE models or frameworks, the primary challenge lies in establishing a uniform input-output schema for diverse tasks. Traditional approaches often involve cumbersome and intricate formats. For instance, some models incorporate examples of In-context Learning(ICL) prior to the main text, enabling the model to discern the appropriate response format. Others integrate Instruction Learning(IL)’s directive text to steer the model’s output. However, these methods not only extend the input text's length, resulting in slower model inference but also fail to address all IE subtasks effectively. In response, our work introduces a streamlined approach. We have developed a unified format capable of accommodating various IE subtasks, significantly reducing input text length without compromising performance, thereby expediting the model’s inference time.

In Figure 3, a variety of input and output examples for specific tasks are presented. The two examples on the left illustrate the TC \& RE and SC \& POS mixed tasks, demonstrating the model's handling of different text-level classification tasks. The process involves encapsulating all TC or SC labels within "<>" mark tokens, signaling to the model both the selection of text-level classification tasks and the initiation and termination points for generating label tokens. Additionally, all ICL samples and IL texts are omitted, retaining only the source text. Notably, the final experimental results indicate that the exclusion of ICL and IL content does not detrimentally impact the model's output.

In the subsequent phase, we address the standardization of input and output formats for text-level labels, and delve into handling word-level labels and their associated entities, spans, and similar elements. As illustrated in Figure 3, our methodology does not incorporate all word-level labels within the input sequence. Instead, we employ the Instruction Word (IW) to direct the model's output. The right side of Figure 3 displays the format of word-level extracted information as produced by the model.

Our approach adapts and enhances the use of the ":" and ";" mark tokens, as borrowed from prior studies, making them suitable for various IE subtasks. In scenarios involving labels and entities, these tokens demarcate the start and end of a label. Specifically, the ":" and ";" tokens precede and follow a label, respectively, leading to the associated entities or spans in behind. When a sentence yields multiple entities, we utilize the capability for multiple overlays, exemplified as ":Label1;Entities1:Label2;Entities2...", to accommodate this complexity.

However, this method is inadequate for tasks involving more complex structures, such as triples or quadruples. To address this, we introduce tailored rules for these specific tasks. For ternary Relationship Extraction (RE) tasks, the ":" token signifies both the start and end of a complete ternary structure, as in ":ObjectLabelSubject:". Conversely, the ";" token is employed to indicate divisions within a ternary, exemplified by "Object;Label;Subject". This innovative approach thus provides a versatile input and output format applicable across all IE subtasks.

\begin{figure}[!ht]
\centering
\includegraphics[width=219 pt]{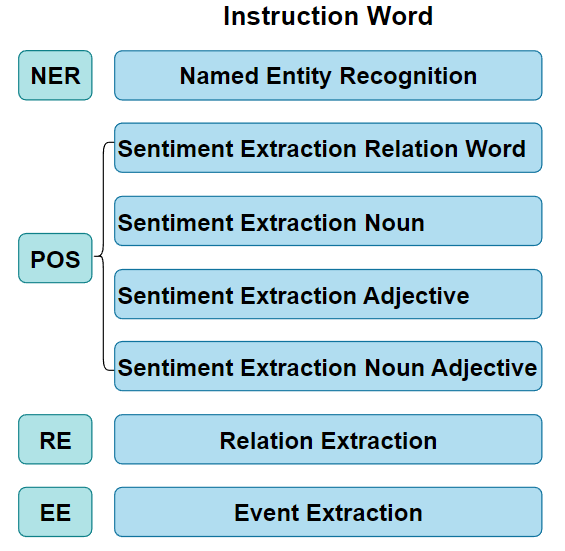}
\caption{\label{4figure4}The illustration depicts Instruction Word in different IE subtasks.}

\end{figure}

\subsection{Instruction Word}\label{Instruction Word}

In this section, we introduce the Instruction Word (IW) methodology. This approach streamlines the process by replacing the conventional lengthy In-Context Learning (ICL) samples and Instruction Learning (IL) directives with concise IWs tailored for specific tasks. These IWs effectively prompt the model to perform word-level information extraction. 

As illustrated on the left side of Figures \ref{3figure3} and Figures \ref{4figure4}, we append distinct IWs to the end of all input sequences. Specifically, the IWs for NER, RE, and EE are "Named Entity Recognition", "Relation Extraction", and "Event Extraction", respectively. A noteworthy aspect is the handling of the Sentiment Classification Part-of-Speech (SCPOS) mix dataset, which is segmented into four distinct sub-datasets. In this case, we uniformly begin with the prefix "Sentiment Extraction" for all four datasets, followed by specific IWs that align with the sub-dataset's focus: these are Relation Word, Noun, Adjective, and Noun Adjective for the respective SCPOS sub-datasets. These tailored IWs effectively guide the model in extracting the comprehensive range of information targeted by the word-level IE subtasks.

\subsection{TCREE Dataset Construction}

In this section, we delineate the process of constructing the Text Classification Relation and Event Extraction (TCREE) dataset. The impetus behind the development of the TCREE dataset stems from a notable gap in the MRE mix dataset. While datasets for NER and sentiment classification are readily available within this mix, representative datasets for the IE domain subtasks—specifically, RE and EE—remain conspicuously absent. The creation of the TCREE dataset, therefore, serves a dual purpose: it not only completes the MRE mix dataset by filling this critical void but also substantiates the presence and significance of MRE within the realms of RE and EE.

\begin{figure}[!t]
\centering
\includegraphics[width=219 pt]{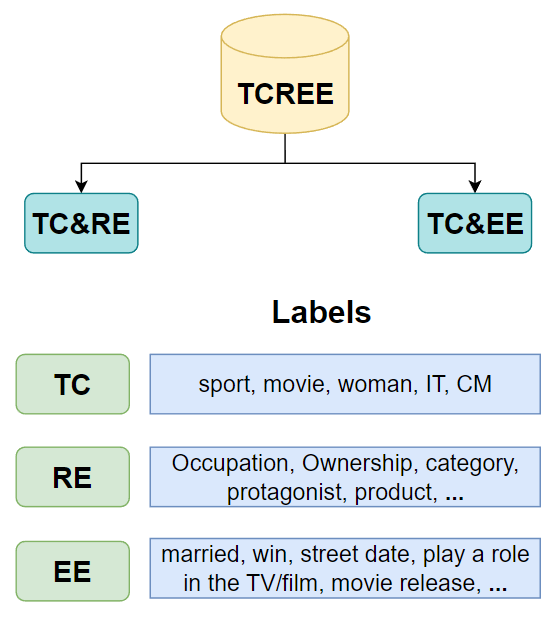}
\caption{\label{5figure5}The illustration depicts Labels of the TCREE dataset.}

\end{figure}

The TCREE dataset is bifurcated into two task categories: TC \& RE (Text Categorization and Relation Extraction) and TC \& EE (Text Categorization and Event Extraction), as depicted in Figure \ref{5figure5}. For any given sentence, it will be annotated with either a single RE ternary relationship or an EE quaternary event, but not both. Moreover, a sentence will not possess more than one instance of an RE ternary or an EE quaternary. In cases where a sentence includes both RE triples and EE quaternions, they are treated as distinct samples and labeled separately.

Figure 5 also provides examples of specific labels for the three tasks. The TC task is limited to news categories, encompassing a predefined set of five topics: sports, movies, women, IT, and CM. Conversely, the RE and EE tasks encompass a broader range of labels, which vary according to the relationships or events depicted in the sentences. Here, we provide a selection of these labels. By excluding less frequently occurring labels, the total count for RE labels stands at seven, while EE labels number eleven.

In addition to specific annotations, the dataset construction process involved the Japanese language Livedoor News Corpus\footnote{\raggedright\url{https://www.rondhuit.com/download.html\#ldcc}} as its foundational corpus. Given the challenge of identifying texts with RE triples or EE quaternions in these original news articles, our approach was twofold. Initially, we performed rule-based extraction on 4851 original news articles, yielding 13,322 texts from a pool of 30,000 to 40,000. From these, 2000 texts featuring the designated RE triples or EE quaternions were manually selected. Subsequently, 1000 of these texts underwent manual annotation. These manually annotated texts formed the training set used to fine-tune the GPT-3.5-Turbo model. The following 1000 texts were then automatically labeled using the fine-tuned GPT-3.5-Turbo. This automatic labeling was subsequently reviewed and manually corrected to ensure accuracy and consistency.

\begin{figure}[!h]
\centering
\includegraphics[width=219 pt]{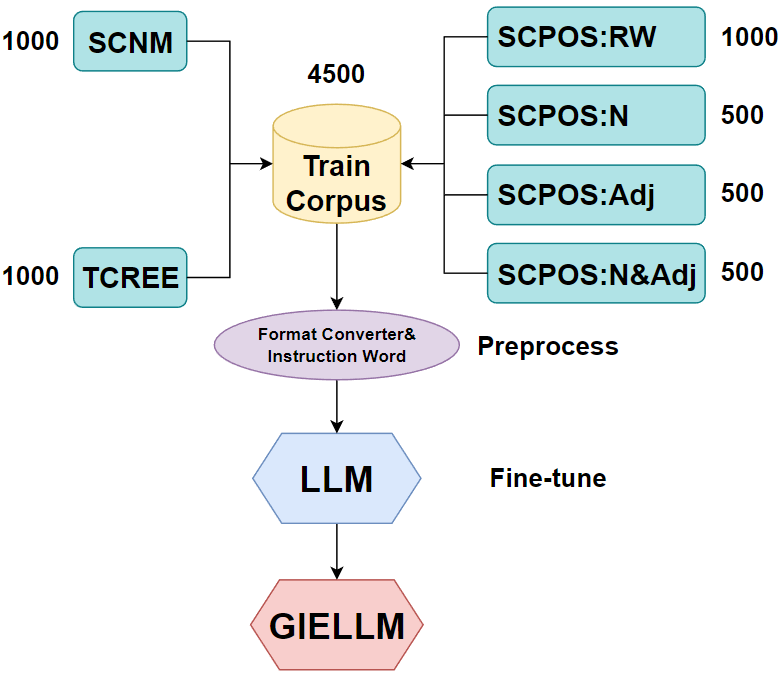}
\caption{\label{6figure6}Detailed Composition of the Mixed Corpus Employed in Training the GIELLM. The numeral adjacent to each dataset indicates its respective size.}

\end{figure}

\section{Experiment Setup}\label{Evaluation}

In this section, we provide a comprehensive explanation of the training process for GIELLM, including the specific parameters and methodologies employed. Additionally, we outline the evaluation metrics used to assess its performance. And you can find the detail of datasets in \ref{sec:datasetdeetailsl}

Figure \ref{6figure6} illustrates the composition of the training corpus, which encompasses a total of 4500 sequences. This corpus undergoes an initial preprocessing phase, which includes formatting conversion and the integration of IW augmentation. The processed corpus is subsequently utilized to fine-tune various LLMs. Through this process, we derive the GIELLM. For our study, we have selected three foundational LLMs: the 7-billion (7B) and 13-billion (13B) parameter models from the LLaMA2\citep{touvron2023llama2} suite, as well as the LLM-jp\footnote{\raggedright\url{https://huggingface.co/llm-jp/llm-jp-13b-v1.0}}, which is pre-trained on the combined pre-training of Japanese and English datasets.

\subsection{Evaluation}

\begin{table*}[ht]
\centering

\begin{tabularx}{1\textwidth}{@{}l*{9}{>{\centering\arraybackslash}X}@{}}
\toprule[2pt]
 & \multicolumn{3}{c}{SCNM} & \multicolumn{3}{c}{SCPOS: RW} & \multicolumn{3}{c}{SCPOS: Adj \& N} \\
 Accuracy &  TL &  WL &  ALL  & TL &  WL &  ALL& TL &  WL &  ALL\\
\midrule
 SLG Framework& \textbf{88.89} & \textbf{81.96} & \textbf{72.41} & 88.21 & 55.57 & 17.28 & 87.30 & 26.22 & 1.60 \\
 USA-7B & - & - & - & \textbf{89.60} & 56.32 & 18.10 & 90.20 & 60.09 & 3.97 \\
 GPT-3.5-Turbo & 49.46 & 11.87 & 6.97 & 53.60 & 14.99 & 1.60 & 73.20 & 10.34 & 0.13 \\
 GIELLM-7B & 85.70 & 63.16 & 54.29 & 86.31 & 66.90 & 25.37 & 92.27 & 54.23 & 3.23 \\
 GIELLM-13B & 85.06 & 54.06 & 45.96 & 83.33 & 65.25 & 24.75 & 90.43 & \textbf{63.98} & \textbf{6.27} \\
 GIELLM-13B-JP & 86.44 & 62.95 & 54.43 & 85.83 & \textbf{67.62} & \textbf{26.15} & \textbf{93.2} & 48.04 & 3.43 \\
\bottomrule[2pt]
\end{tabularx}
\vspace{1em}

\begin{tabularx}{1\textwidth}{@{}l*{9}{>{\centering\arraybackslash}X}@{}}
\toprule[2pt]
\toprule[2pt]
 & \multicolumn{3}{c}{SCPOS: N} & \multicolumn{3}{c}{SCPOS: Adj} & \multicolumn{3}{c}{TCREE} \\
 Accuracy &  TL &  WL &  ALL  & TL &  WL &  ALL& TL &  WL &  ALL\\
\midrule
 SLG Framework& 89.50 & 27.62 & 3.00 & 83.00 & 73.84 & 52.47 & 96.49 & 73.12 & 71.72 \\
 USA-7B & 91.50 & 62.41 & 6.86 & 92.17 & 64.94 & 50.90 & - & - & - \\
 GPT-3.5-Turbo & 73.83 & 10.44 & 0.23 & 78.83 & 15.45 & 9.87 & 80.02 & 7.33 & 6.73 \\
 GIELLM-7B & 92.10 & 58.33 & 4.60 & 91.6 & 75.71 & 58.8 & 97.19 & 77.01 & 75.80 \\
 GIELLM-13B & 90.13 & \textbf{68.28} & \textbf{9.80} & 91.9 & 77.58 & 60.70 & 94.58 & 74.90 & 73.29 \\
 GIELLM-13B-JP & \textbf{92.43} & 49.63 & 4.43 & \textbf{93.47} & \textbf{78.71} & \textbf{63.33} & \textbf{97.49} & \textbf{78.51} & \textbf{77.41} \\
\bottomrule[2pt]
\end{tabularx}
\vspace{1em}

\caption{\label{table1results}
The accuracy of Six Mix Datasets in different Method or Models. SLG Framework use fine-tune method. USA model use 1-shot ICL and IL methods. GPT-3.5-Turbo use 1/5-shot and IL methods. GIELLM model use Format Converter and IW methods(No ICL and IL).}
\end{table*}

In this study, accuracy serves as the uniform evaluation metric for all experiments. Our methodology involves calculating accuracy at both word-level and text-level sequences independently. For word-level accuracy, we adopt the following procedure: Initially, label-span pairs from the predicted and actual sequences are extracted, denoted as \( P_{\text{pair}} \)
 and \( A_{\text{pair}} \), respectively. We then count the number of matching label-span pairs \( M_{\text{match}} \) between \( P_{\text{pair}} \) and \( A_{\text{pair}} \). Finally divided by the total number of label-span pairs (\( T_{\text{pair}} \)) in the actual sequence as the final \( WL_{\text{accuracy}} \). The word-level accuracy (\( WL_{\text{accuracy}} \)) is calculated using the formula:
\begin{equation}
     WL_{\text{accuracy}}  = \frac{M_{\text{match}}}{T_{\text{pair}}}
\end{equation}
In contrast, the calculation of text-level accuracy (\( TL_{\text{accuracy}} \)) is more straightforward. We directly extract and compare the labels enclosed within "<>" from both the predicted and actual sequences. \( TL_{\text{accuracy}} \) is determined by the ratio of the correctly predicted labels to the total number of labels \( T_{\text{label}} \), as follows:
\begin{equation}
     TL_{\text{accuracy}}  = \frac{M_{\text{match}}}{T_{\text{label}}}
\end{equation}
When the \( WL_{\text{accuracy}} \) and \( TL_{\text{accuracy}} \) in a sequence are both \(100\%\), we count the sequence as having an overall accuracy of \(100\%\). Finally, the number of sequences with \(100\%\) overall accuracy is divided by the total number of sequences to calculate \( ALL_{\text{accuracy}} \).

Regarding the datasets, due to constraints in resources and time, we employed a unique approach for the three SCPOS subsidiary datasets. For these, we randomly selected samples of 1000 for three separate test sets, averaging the accuracies obtained from these three iterations to represent the final accuracy measure. Conversely, for the other three datasets, we utilized their entirety as the test set.

\section{Results}

To establish a baseline for comparison, we analyzed various frameworks and models from previous studies, focusing on their application to mixed datasets. The Sentence-to-Label Generation Framework (SLG Framework) stands out for its capability to process both word-level and text-level tasks, employing the T5 model as its core. In contrast, the USA-7B, a large language model (LLM) specifically designed for sentiment analysis, is limited to handling only four sub-datasets of SCPOS. Additionally, we included GPT-3.5-Turbo in our baseline assessment, applying IL and ICL for evaluation.

As indicated in Table \ref{table1results}, the GIELLM model sets new benchmarks in 14 out of 18 accuracy metrics across six datasets, showcasing its remarkable effectiveness. Notably, in the SCNM dataset, where GIELLM does not secure the top position, its performance still significantly surpasses that of GPT-3.5-Turbo. In the relatively straightforward TC dataset, GIELLM achieves an accuracy of 86.44, closely approaching the 88.89 accuracy of the finely-tuned SLG Framework. Most impressively, GIELLM outstrips the task-specific, fine-tuned SLG Framework across the remaining five datasets, underscoring its robust generalization capabilities in IE subtasks.

Further analysis reveals that the bilingual model, GIELLM-13B-JP, which is pre-trained in both Japanese and English, generally outperforms its monolingual\footnote{The LLaMA2 Model is mainly pre-trained on the English Corpus. And GIELLM-13B is fine-tuned from LLaMA2} counterpart, GIELLM-13B. This finding is particularly significant as it also surpasses other predominantly English-trained LLMs like LLaMA2 by a substantial margin. Interestingly, the results suggest that model performance does not linearly correlate with LLM size; in several instances, GIELLM-7B achieves higher accuracy than GIELLM-13B. This observation warrants additional exploration in future research to better understand the intricacies of LLM performance relative to their size in the IE subtasks.

\section{Analysis MRE in TCREE task}

In our study, we utilized the SLG Framework, based on the T5-base model, to conduct the MRE)ablation experiment on the newly developed TCREE dataset. The selection of the SLG Framework was strategic, aiming to maintain strict control over all variables except the dataset under examination. Consequently, we opted not to employ LLMs for this test. The SLG Framework was chosen for its capability to not only handle the TCREE dataset but also to effectively process mixed datasets, as well as separate word-level and text-level tasks.

For the purpose of this study, we segregated the Text Classification (TC) task and the Relation \& Event Extraction (REE) task into two distinct datasets and conducted tests on each. As indicated in Table \ref{table2mre}, the accuracies observed in both the TC and REE tasks on these separate datasets exhibited a notable decline when compared to their performances on the mixed dataset. This outcome further corroborates the existence of a Mutual Reinforcement Effect between the TC and REE tasks. This also proves once again that Mutual Reinforcement Effect between in TC and REE tasks. The MRE makes the final performance 1 + 1 > 2.

\begin{table}[!t]
\centering
\begin{tabular}{lccc}
\hline
  &  & \textbf{Accuracy} &   \\
  \textbf{Dataset}  & \textbf{TL} & \textbf{WL} &  \textbf{ALL} \\
\hline
TCREE & \textbf{96.49} & \textbf{73.12} & \textbf{71.72} \\
Single TC & 95.89 & - & - \\
Single REE & - & 58.73 & - \\

\hline
\end{tabular}
\caption{\label{table2mre}
 SLG-Framework(T5-base)}
\end{table}

\section{Conclusion}

For an extended period, individual IE subtasks have been managed by dedicated, task-specific fine-tuned models. The advent of LLMs heralds the potential for consolidating all IE subtasks under a single LLM framework. In this paper, the last piece of the puzzle of MRE mix datasets is completed by constructing the TCREE dataset. Concurrently, we introduce GIELLM, a unified model designed to encompass all principal IE subtasks. The exceptional performance of GIELLM suggests a promising future where a single LLM could potentially replace a suite of ten narrowly focused, fine-tuned models. Due to resource and time limitations, our current demonstration is confined to Japanese. Nonetheless, \textbf{we posit that the MRE-based GIELLM model will catalyze a paradigm shift for researchers in the domain of IE.}

\bibliography{custom}

\appendix

\section{Dataset Details}
\label{sec:datasetdeetailsl}

\begin{table}[!h]
\centering
\begin{tabular}{lll}
\hline
  & \textbf{train set} & \textbf{test set}  \\
\hline
 SCNM & 1000 & 4343\\
 SCPOS:RW& 1000 & 1000 \\
 SCPOS:N& 500 & 187028  \\
 SCPOS:Adj& 500 & 187028  \\
 SCPOS:N \& Adj& 500 & 187028  \\
 TCREE & 1000 & 1000  \\
\hline
\end{tabular}
\caption{\label{table3mixdataset}
The table provides statistical data, delineating the total counts of training and testing sets for six Mutual Reinforcement Effect (MRE) based Japanese datasets. SCNM: Sentence Classification and Named Entity Recognition Mix Dataset. SCPOS: Sentiment Classification and Part-of-Speech Dataset. RW: Relation Word. N: Noun. Adj: Adjective. N \& Adj: Nous and Adjective. TCREE: Text Classification and Relation \& Event Extraction Dataset.}
\end{table}

\textbf{Datasets.} The initial phase involved dataset selection. We selected the SCNM (Sentence Classification and Named Entity Recognition Mix Dataset) and the SCPOS (Semtiment Classification and Part-of-Speech Dataset) dataset, which comprises four sub-datasets, in addition to the TCREE dataset tagged in this study. In total, there are six datasets, each partitioned into training and testing sets. It is important to note that in the original text, the Adjective sub-dataset of SCPOS was labeled 'Verb \& Adjective.' In this case, we have reclassified the verbal adjective solely as an adjective, excluding the verb. Similarly, the original 'Sentiment Relation Word' (SRW) dataset was condensed to 'Relation Word' (RW), following the same approach used in the prior IW design. As detailed in Table \ref{table3mixdataset}, we randomly divided these six datasets into training and testing sets. Importantly, when determining the size of the training set, we adhered to the principle outlined in the LIMA paper\citep{zhou2023lima}, emphasizing dataset quality over quantity. Accordingly, we selected 1000 sequences for each of the three primary datasets and a smaller number, 500, for each of the SCPOS subsidiary datasets.

\end{document}